\documentclass[conference,a4paper]{APSIPA2021}
\usepackage{amsmath}
\usepackage{amsfonts}
\usepackage{graphicx}
\usepackage{multirow}
\usepackage{hhline}
\usepackage{hyperref}
\usepackage{threeparttable}
\usepackage[backend=biber,style=ieee,]{biblatex}
\usepackage{geometry}
\usepackage{fancyhdr}

\fancypagestyle{firststyle}{
  \fancyhf{}
  \fancyhead[C]{2023 Asia Pacific Signal and Information Processing Association Annual Summit and Conference (APSIPA ASC)}
}

\begin{document}

\title{Communication-Efficient Design of Learning System for Energy Demand Forecasting
of Electrical Vehicles\thanks{Work supported by NSF CCF grants 2007527 and 2007754.}}

\author{
\authorblockN{
Jiacong Xu\authorrefmark{1}, Riley Kilfoyle\authorrefmark{2}, 
Zixiang Xiong\authorrefmark{2} and Ligang Lu\authorrefmark{3}
}

\authorblockA{
\authorrefmark{1}Dept of CS, Johns Hopkins University, Baltimore, MD 21218}

\authorblockA{
\authorrefmark{2}Dept of ECE, Texas A\&M University, College Station, TX 77843}

\authorblockA{
\authorrefmark{3}Shell International Exploration and Production Inc., Houston, TX 77082\\
jxu155@jhu.edu; rskilfoyle@tamu.edu; zx@ece.tamu.edu; ligang.lu@shell.com
}
}

\maketitle
\thispagestyle{firststyle}
\pagestyle{fancy}

\begin{abstract}
Machine learning (ML) applications to time series energy utilization forecasting problems are a challenging assignment due to a variety of factors. Chief among these is the non-homogeneity of the energy utilization datasets and the geographical dispersion of energy consumers. Furthermore, these ML models require vast amounts of training data and communications overhead in order to develop an effective model. In this paper, we propose a communication-efficient time series forecasting model combining the most recent advancements in transformer architectures implemented across a geographically dispersed series of EV charging stations and an efficient variant of federated learning (FL) to enable distributed training. The time series prediction performance and communication overhead cost of our FL are compared against their counterpart models and shown to have parity in performance while consuming significantly lower data rates during training. Additionally, the comparison is made across EV charging as well as other time series datasets to demonstrate the flexibility of our proposed model in generalized time series prediction beyond energy demand. The source code for this work is available at \url{https://github.com/XuJiacong/LoGTST_PSGF}.
\end{abstract}

\section{Introduction}
\let\thefootnote\relax\footnotetext{Work supported in part by NSF grants ECCS-1923803, CCF-2007527, CCF-2324397, and Shell.}
The Alternative Fuels Data Centre lists more than 54,000 EV charging stations (CSs) currently in operation in the United States. It is projected that the number of EVs on the road will rise by more than 4000\% by 2030 \cite{2030}. As a result, there is a strong need to effectively and intelligently predict energy demands for EV CSs so as to mitigate their impact on the power grids without upgrading/expansion capability. Generally, the power grid supplies the energy for CSs once requests from EVs are received. Predicting the amount of energy needed at each CS across different periods of time will allow the power suppliers of CSs to purchase the desired amount of electricity at lower rates in order to save money and make charging at public stations more efficient – a key step towards smart charging \cite{keynote}. In addition, the power grid can coordinate energy consumption via schedule management to reduce energy costs and waste of resources.

Energy utilization is a form of time series prediction algorithm that has been thoroughly studied in the past. Majidpour et al. \cite{Maj} compared fast machine learning-based time-series prediction algorithms and found that the nearest neighbor algorithm showed improved accuracy. Ryu et al. \cite{kim} proposed deep learning (DL) load forecasting models and showed that DL methods exhibited better performance compared to other forecasting models. Paterakis et al. \cite{van} compared a DL method with the eight most commonly used machine learning methods such as nearest neighbors, support vector machines, Gaussian processes, and regression trees, in time series energy prediction and showed that the DL method outperformed all eight other methods.
However, conventional DL algorithms trained on individual CS nodes may not achieve high prediction accuracy due to insufficient data and failure to consider the influence of other CSs, particularly the ones in the neighborhood \cite{ho}.

We introduce machine learning-based approaches that can not only improve the accuracy of time series prediction but also significantly reduce the communication overhead involved in training a machine learning algorithm involving a dispersed series of nodes. In particular, we first introduce a communication model using the power provider as a centralized node to gather all information from the individual CS nodes in a considered metropolitan area (e.g., Houston, or London, or Amsterdam, etc.). We then develop an ML algorithm to help the power provider more accurately predict energy demands for the CSs while simultaneously utilizing less data than previous algorithms. Our focus thus is two-fold. First to more accurately predict energy usage at CSs and second to lower the data utilization required to implement the prediction algorithm across a broad collection of CSs spread out geographically.

\begin{figure}[tbh]
\begin{center}
\includegraphics[width=90mm]{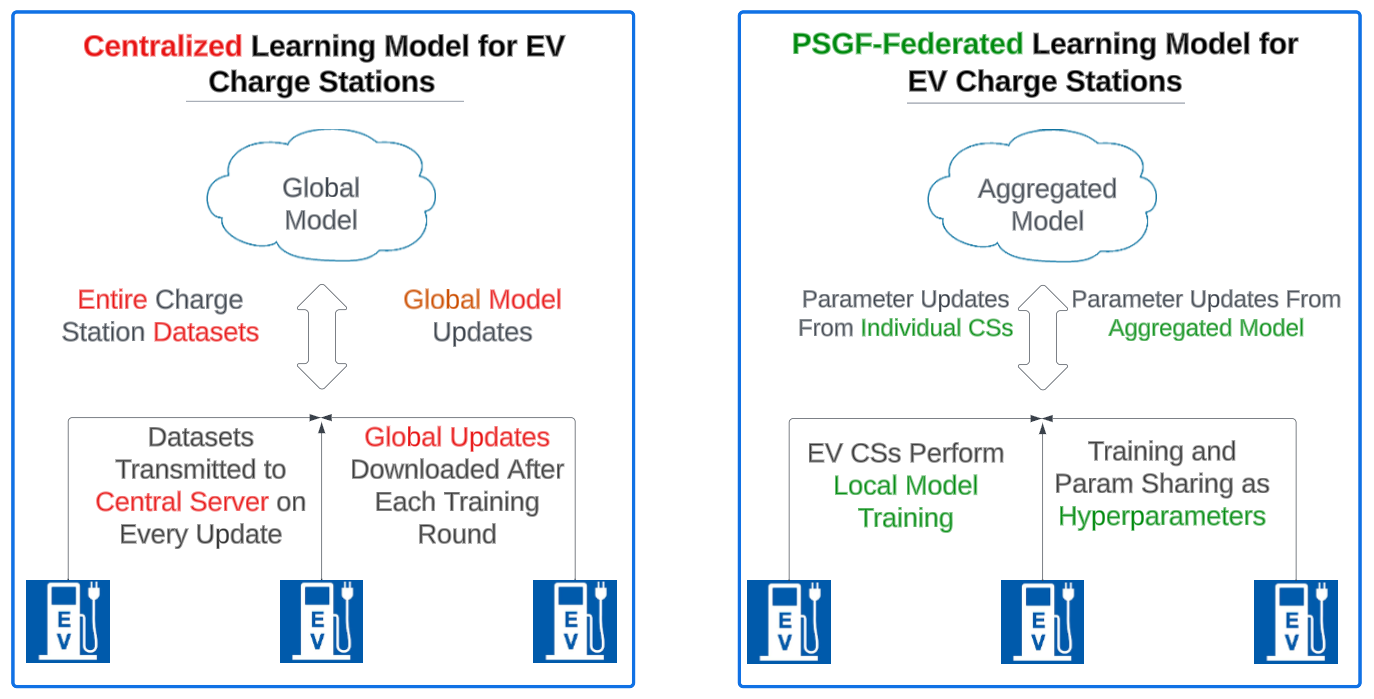}
\end{center}
\vspace*{-.2in}
\caption{FL architectures vs Centralized: In centralized learning (left), data is sent to the cloud, where the ML model is built. The model is accessed by a user through an API sending a request to access one of the available services.  In FL [7,8] (right), each CS trains a model and sends only its updated parameters to a server for aggregation. Data is kept locally on devices and knowledge is shared selectively with peers through an aggregated model.}
\label{FLfigure}
\end{figure}

Specifically, we adopt a federated learning (FL) approach \cite{li,survey} to energy demand prediction. 
As seen in Fig. \ref{FLfigure}, FL is a form of distributed learning wherein updates to model parameters are calculated locally on individual devices rather than transmitting massive volumes of data to centralized servers for global model development. 
These local model updates are then transmitted
to a centralized server for aggregation into a global model.
Once the global model has been updated according to various
algorithms, the updated global model is transmitted out to the
CSs for the next round of training. In this fashion
the individual data never leaves the local CSs, some of the
computation is offloaded to local CSs, and a globally useful
DL model is created. This reduces the increasing strain placed on
communication networks while accomplishing the goal of training
a useful global model. Thus, the advantage with FL is that the CSs only need to share their trained model parameters obtained from their individual local datasets instead of sharing their entire datasets, which generally are of large volume and often are protected by data privacy laws and security regulations.  

\vspace*{-.1in}

%\begin{figure}[tbh]
%\begin{center}
%\includegraphics[width=85mm]{Latex-2023-0511/Dundee.png}
%\end{center}
%\caption{Centralized vs FL architectures: In centralized learning (left), data is sent to the cloud, where ML model is built. The model is used by a user through an API by sending a request to access one of the available services.  In FL \cite{li,survey} (right), each device trains a model and sends its parameters to the server for aggregation. Data is kept on-devices and knowledge is shared through an aggregated model with peers.}
%\label{Dundee}
%\end{figure}

The EV charging prediction task for an individual client is a uni-variate time-series forecasting problem, which has been previously investigated. For example,
using the UK EV dataset \cite{dun}, Saputra et al. \cite{1ref} showed that simple location-based node clustering improves the accuracy of energy demand prediction up to 24.63\% and decreases communication overhead by 83.4\% compared with other baseline machine learning algorithms. 
Different from previous works \cite{1ref,ho} that treat the hourly energy consumption as the prediction output, we aggregate 24 hours of energy consumption as one data point upon which to perform daily prediction. This time aggregation is done to assist in dealing with inconsistencies present in various portions of the UK EV dataset \cite{dun} wherein certain chargers were offline for maintenance etc. at random times and therefore present discontinuities on an hourly level, whereas the daily data exhibits more clear trends of energy consumption variation and less meaningless randomness.

Inspired by the great success of transformer architectures for language and vision tasks, many advanced transformer architectures have been proposed for time series forecasting in the last two years. These transformer-based models have demonstrated better prediction performance compared with traditional RNNs and recent work PatchTST \cite{Yuqietal-2023-PatchTST} outperforms its multi-layer perceptron (MLP) counterpart and shows state-of-the-art performance when tested on standard time-series forecasting benchmark datasets. In this paper, we incorporate the ideas from recent advances in efficient 2-D vision transformers and propose an improved lighter model called Local and then Global Time Series Transformer (LoGTST).

For generic FL, sharing model parameters among all clients and the server will involve heavy communications overhead costs which has the unwanted byproduct of slower convergence in real-time. To reduce the communication cost and convergence time, Online-Fed randomly down-selects a specific set of CS clients for model updating. Partial Sharing-based Online FL (PSO-Fed) \cite{gogineni2022communication} further alleviates this problem by only randomly sharing a partial subset of model parameters to those selected clients. However, in each global iteration, the clients in both of these two models can only access their own information to train on or stay idled not training during this particular iteration, which will hinder the convergence speed and limit that specific CS's generalization ability. Thus, we propose Partial Sharing Global Forwarding FL (PSGF-Fed) that is built upon PSO-Fed but enables the server to randomly share a small number of partial parameters to all the clients thus enabling all CSs to train during each iteration. In this way, PSGF-Fed is able to reduce the total communication overhead by accelerating the convergence speed of each local model so that fewer total iterations of training are required.

Through the generated results, we demonstrate that our proposed novel LoGTST architecture can achieve similar prediction performance with current baseline model: PatchTST \cite{Yuqietal-2023-PatchTST} but involves around half of its number of trainable parameters, and our PSGF-Fed updating strategy offers better trade-off between prediction accuracy and communication cost compared with PSO-Fed \cite{gogineni2022communication} and Online-Fed. Also, we present additionally the flexibility via hyper-parameters between model performance and data overhead. This serves to demonstrate that our combination of LoGTST prediction model and PSGF-Fed distributed training strategy constitutes an efficient ML pipeline for energy demand forecasting of electrical vehicles.

\section{Method}
The learning system in this application consists of two major parts: the DL model and the FL strategy. For better prediction accuracy, the DL model should be able to capture the local and global trends of the input time series, and the FL policy should aggregate the information from all the clients and lead the model to a quick convergence with minimal bandwidth consumption. With the constraints of communication overhead driving larger parameter sets, the space complexity of the learning model should be as small as possible and the time and the amount of information sharing between server and clients should be minimized.

\subsection{MetaFormer}

Current research on the model architecture design for time series forecasting can be divided into two camps: Transformer and  MLP. Different from previous transformer architectures, PatchTST \cite{Yuqietal-2023-PatchTST} splits a time-series into patches and tokenizes these patches into vectors by patch embedding following Vision Transformer (ViT) \cite{dosovitskiy2021an}. This new transformer outperforms MLPs and other transformers such as DLinear \cite{zeng2023transformers} and FEDFormer \cite{zhou2022fedformer} on many popular benchmarks.

The self-attention operation \cite{vaswani2017attention} in traditional transformers possesses the advantages of global dependency parsing and dynamic weight generation. However, self-attention brings a large amount of space and time complexity to the overall model. The authors of MetaFormer \cite{yu2022metaformer} summarize recent advances on vision transformer and argue that it is the transformer architecture itself that contributes to the performance of ViT on vision tasks. They also replace the self-attention operation with a simple pooling operation and demonstrate that PoolFormer shows better efficiency than most recent variants of ViT considering the complexity and the accuracy.

\begin{figure}[tbh]
\begin{center}
\includegraphics[width=85mm]{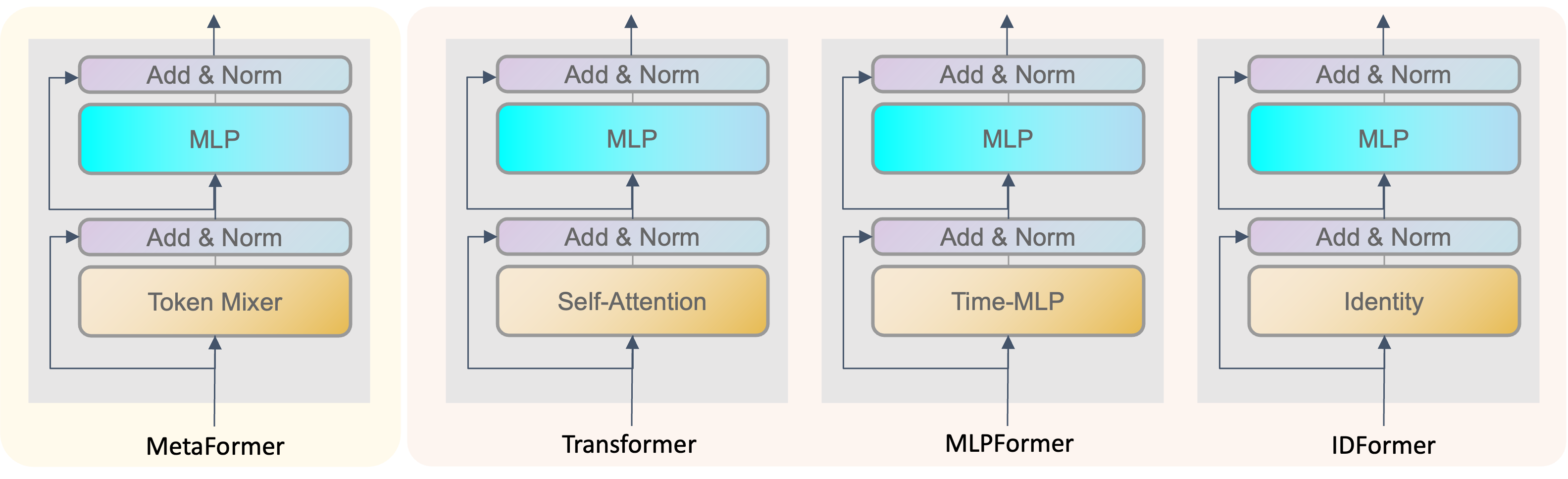}
\end{center}
\vspace*{-.2in}
\caption{Architectures of MetaFormer and its variants. Time-MLP refers to the MLP operation along the series of tokens; Identity means there is no operation.}
\label{metaformer}
\end{figure}

Similar to PatchTST, we introduce the MetaFormer architecture into the task of 1-D time series forecast. As shown in Figure \ref{metaformer}, we replace self-attention with Time-MLP and Identity operations as the token-mixer and propose two variants of MetaFormer: MLPFormer and IDFormer. In our experiments, we observed the surprising finding that the simplest identity operation performs as well as PatchTST. 
Since our goal is to reduce the model complexity while minimizing loss of accuracy, we favor most of the transformer blocks to be the simplest IDFormer in our final model.

\vspace*{-.1in}

\subsection{LoGTST}
\begin{figure*}[tbh]
\begin{center}
\includegraphics[width=160mm]{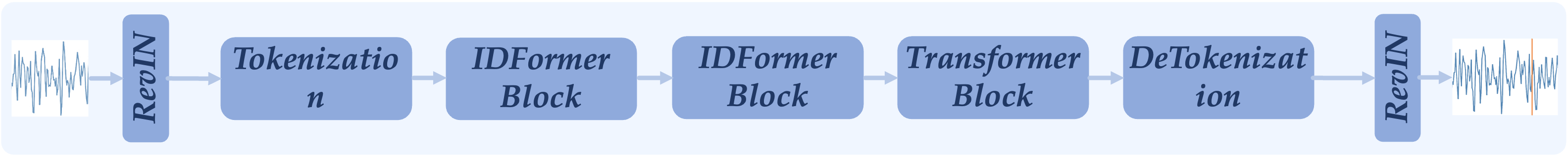}
\vspace*{-.2in}
\end{center}
\caption{Overview of the architecture of our proposed \textbf{LoGTST}. The input and output signals are processed by Reversible Instance Normalization (RevIN) module before and after the learning model, respectively. For fair comparison with PatchTST, we only change the first and second transformer blocks to IDFormer blocks so that the model can fully process the local features and keep the final transformer blocks for parsing of global dependency.}
\label{model}
\end{figure*}

Figure \ref{model} depicts the basic architecture of our proposed LoGTST model. 

\textbf{RevIN}. The Reversible Instance Normalization (RevIN) \cite{kim2021reversible} module normalizes the input signal for each sample, which consists of one look-back window and one prediction horizon, and records the normalization factors for future denormalization of the output signal. In this way, RevIN is able to symmetrically remove and restore the statistical information of a time-series instance and improve the model’s stability.

\textbf{Tokenization and DeTokenization}. The tokenization process is the same as Patch Embedding in PatchTST \cite{Yuqietal-2023-PatchTST}. Here we call it Tokenization to follow the convention in computer vision and natural language processing. Given the input uni-variate time series as $\textbf{x}$ and its length as $L$, the tokenization process can be simply accomplished by applying 1-D convolution on $\textbf{x}$ with a predefined kernel of size $P$ and stride $S$. The kernel size can also be seen as the patch length and the number of tokens is $N=[L/S]$. The DeTokenization process directly flattens feature vectors $\textbf{x}^{(i)}_{hidden}$, where $i=1,2,...,N$, concatenates the flattened vectors, and applies MLP for prediction, which can be represented as
\begin{equation}
    Pred = MLP\{Concat[Flat(\textbf{V}_0), Flat(\textbf{V}_1), ...]\}
\end{equation}

\textbf{IDFormer and Transformer}: In the transformer branch, we replace the early two transformer blocks by IDFormer blocks to reduce total model complexity and prevent early attention on naive features and keep the last transformer block to parse dependencies between hidden vectors. The patch embeddings $\textbf{x}_p \in \mathbb{R}^{D \times N}$ should be merged with additive learnable positional encoding $\textbf{W}_{pos}$ by $\textbf{x}_{d} = \textbf{x}_p + \textbf{W}_{pos}$ in case of the attention mechanism treating all the feature vectors equally. For each head $h=1,2,...,H$, we define three learnable matrices: $W^{Q}_h, W^{K}_h \in \mathbb{R}^{D \times d_k}$ and $W^{V}_h \in \mathbb{R}^{D \times D}$. Then, the calculation of multi-head self-attention can be written as:
\begin{equation}
    O^{T}_h = Attention(Q_h, K_h, V_h) = softmax\{\frac{Q_hK^{T}_h}{\sqrt{d_k}}\}V_h
\end{equation}

\textbf{Loss}: Following previous work for time series forecasting \cite{zhou2022fedformer, Yuqietal-2023-PatchTST}, we also use MSE to measure the discrepancy between the prediction and the reality. Defining the number of input variables as $M$ and the prediction horizon as $T$, then the total loss can be calculated by $\mathcal{L}=1/M \sum^M_{i=1}\Vert \hat{\textbf{x}}^{(i)}_{L+1:L+T}-\textbf{x}^{(i)}_{L+1:L+T} \Vert^2$. Note that there is only one channel ($M=1$) for the task EV charging forecasting.

\subsection{PSGF-Fed}
\begin{figure*}[t]
\begin{center}
\includegraphics[width=160mm]{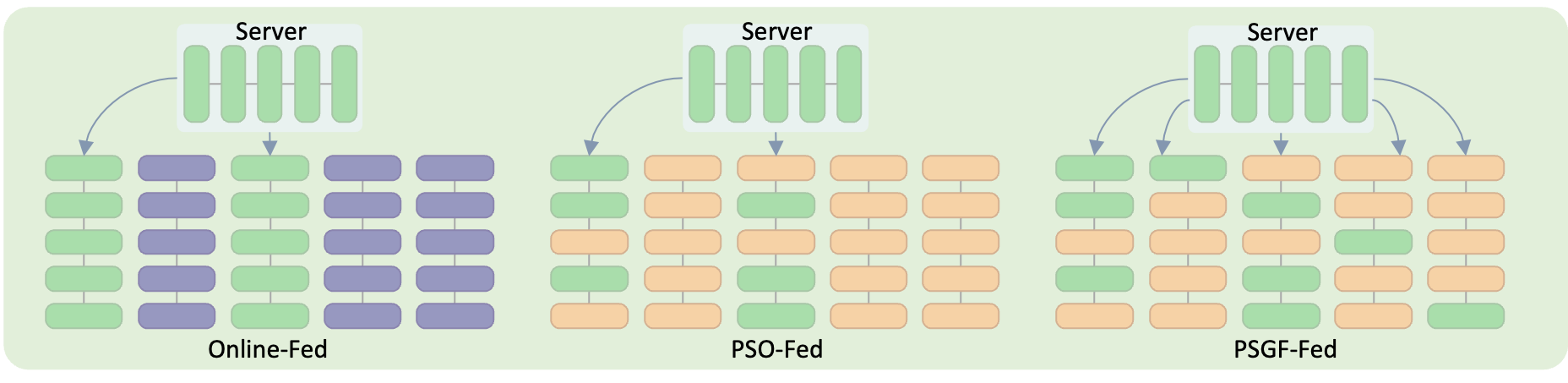}
\vspace*{-.2in}
\end{center}
\caption{Illustration of the model parameters' sharing from server to clients for Online-Fed, PSO-Fed \cite{gogineni2022communication}, PSGF-Fed (ours). The green rectangles refer to shared parameters. The purple rectangles will remain idle during the following local update while the orange and green rectangles will be updated. }
\label{fed}
\end{figure*}

The difference between our proposed PSGF-Fed and previous works is illustrated in Figure \ref{fed}.

\textbf{Online-Fed}. Instead of exchanging the model parameters with all the clients, the server of Online FL (Online-Fed) will randomly select a subset of clients in every iteration for communication efficiency. Denote the selected subset of client indices as $S_n$, where $C=|S_n|$ refers to the number of selected clients and $n$ represents the current iteration. There is no need to perform local updates for unselected clients because the local model will be directly replaced by the server model when these clients are selected in future iterations. Define the local model parameters for client $i$ in iteration $n$ after local update as $\textbf{w}^{i}_{n+1}$, then the global model can be updated by:
\begin{equation}
    \textbf{w}_{n+1}=\frac{1}{C}\sum_{i \in S_n}\textbf{w}^{i}_{n+1}
\end{equation}

\textbf{PSO-Fed} \cite{gogineni2022communication}. It reduces the granularity of communication to the parameter level for better efficiency and randomness. PSO-Fed randomly selects a subset of clients $S_n$ and also randomly selects a subset of parameters $\textbf{S}^{i}_n$ for client $i$ for parameter exchange between server and clients. Different from Online-Fed, the unselected clients can still update their local models because when they are selected, not all of the local parameters will be replaced. The $\textbf{S}^{i}_n$ can be a $D \times D$ diagonal matrix with $M$ ones for selected diagonal elements and $D-M$ zeros. Then the selected local model can be updated by:
\begin{equation} \label{lc_up}
    \textbf{w}^{i}_{n+1}=LocalUpdate(\textbf{S}^{i}_n\textbf{w}_{n}+(\textbf{I}_D-\textbf{S}^{i}_n)\textbf{w}^i_{n})
\end{equation}
The server generates the global model by aggregating the local parameters and this can be accomplished by:
\begin{equation}
    \textbf{w}_{n+1}=\frac{1}{C}\sum_{i \in S_n}\textbf{S}^{i}_{n+1}\textbf{w}^{i}_{n+1}+(\textbf{I}_D-\textbf{S}^{i}_{n+1})\textbf{w}_{n}
\end{equation}
The communication overhead is then significantly reduced compared with Online-Fed while maintaining comparative convergence speed.

\textbf{PSGF-Fed}. PSO-Fed enables self-learning for each unselected client, which guarantees its fast convergence speed even though only partial parameters are shared. However, training a local model only on limited local data for many local iterations (or even several global iterations) without any global information has the risk of overfitting or weakening the generalization ability. To alleviate this issue, we propose a new online FL policy, namely PSGF-Fed, where the server will randomly select a small subset of model parameters and share them with each client. In this way, all the clients will receive some global information from the server, which will regularize the local training process. Define $\textbf{F}^{i}_n$ as the parameter forwarding diagonal matrix of unselected client $i$ with $N$ ones for diagonal elements (selected forwarding parameters) and $D-N$ zeros, then the local training for this client can be represented by
\begin{equation} \label{un_lc_up}
    \textbf{w}^{i}_{n+1}=LocalUpdate(\textbf{F}^{i}_n\textbf{w}_{n}+(\textbf{I}_D-\textbf{F}^{i}_n)\textbf{w}^i_{n})
\end{equation}
The number of selected forwarding parameters ($N$) for each client can be adjusted to reach the best trade-off between convergence speed and communication overhead. All the selected local models will still be updated by (\ref{lc_up}) following PSO-Fed.

%\section{Experiments}
\section{Experiments}

\subsection{Centralized Time Series Forecasting}
\subsubsection{Datasets} Following previous works, we evaluate our model performance on five public-domain datasets: Weather and four electricity transformer temperature (ETT) datasets (ETTh1, ETTh2, ETTm1, ETTm2) \cite{zhou2021informer}, which contain multivariate time series with more than 10k timesteps. These five datasets have been extensively utilized for benchmarking time series forecasting models and are introduced in detail by \cite{wu2021autoformer}. Since our ultimate goal is to deal with the prediction of uni-variate EV charging task, the forward process for each channel uni-variate series is independent of LoGTST and the model weights are shared between different channels. 

\subsubsection{Setting} For fair comparison, we build our model based on the codebase of PatchTST \cite{Yuqietal-2023-PatchTST}, FEDformer \cite{zhou2022fedformer}, and Autoformer \cite{wu2021autoformer} to guarantee the same experimental setup. Specifically, we train the model by stochastic gradient descent using Adam optimizer \cite{kingma2014adam} and adjust the learning rate according to the cycle learning rate policy \cite{smith2019super}. The total number of epochs is 100, but the training process will be early stopped with patience of 20 epochs to prevent overfitting. We also include previous representative models Informer \cite{zhou2021informer} and Pyraformer \cite{liu2021pyraformer} in our result comparison.

\begin{table*}[tbh]
\caption{Experimental results of LoGTST (ours) and other recently proposed models for centralized time series forecasting task. We calculate Mean Square Error (MSE) and Mean Absolute Error (MAE) of the predicted series and ground truth for all the models to quantify their performance. Following previous works, we also change the prediction length to be ${96, 192, 336, 720}$ to fully examine the long-time series forecasting performance. Note that PatchTST/64 has a longer look-back window than PatchTST/42.}
\centering
\begin{tabular}{c|c|cc|cc|cc|cc|cc|cc|cc} 
\hline\hline
\multicolumn{2}{c|}{Models}         & 
\multicolumn{2}{c|}{LoGTST}   &
\multicolumn{2}{c|}{PatchTST/64 \cite{Yuqietal-2023-PatchTST}} & \multicolumn{2}{c|}{PatchTST/42 \cite{Yuqietal-2023-PatchTST}} & \multicolumn{2}{c|}{FEDformer \cite{zhou2022fedformer}}  & 
\multicolumn{2}{c|}{Autoformer \cite{wu2021autoformer}} & 
\multicolumn{2}{c|}{Informer \cite{zhou2021informer}} & 
\multicolumn{2}{c}{Pyraformer \cite{liu2021pyraformer}}  \\ 
\hline\hline
\multicolumn{2}{c|}{\#Parameters}   & \multicolumn{2}{c|}{\textbf{5.39E+05}} & \multicolumn{2}{c|}{1.19E+06}    & \multicolumn{2}{c|}{9.21E+05}    & \multicolumn{2}{c|}{1.63E+07}    & \multicolumn{2}{c|}{1.05E+07}   & \multicolumn{2}{c|}{1.13E+07} & \multicolumn{2}{c}{1.01E+07}    \\ 
\hline
\multicolumn{2}{c|}{Metric}         & MSE   & MAE                   & MSE   & MAE                      & MSE   & MAE                      & \multicolumn{1}{c|}{MSE} & MAE   & MSE   & MAE                     & MSE   & MAE                   & MSE   & MAE                     \\ 
\hline
\multirow{4}{*}{Weather} & 96       & 0.151 & 0.199                 & \textbf{0.149} & \textbf{0.198}                    & 0.152 & 0.199                    & 0.238                    & 0.314 & 0.249 & 0.329                   & 0.354 & 0.405                 & 0.896 & 0.556                   \\
                         & 192      & 0.195 & \textbf{0.240}                 & \textbf{0.194} & 0.241                    & 0.197 & 0.243                    & 0.275                    & 0.329 & 0.325 & 0.370                   & 0.419 & 0.434                 & 0.622 & 0.624                   \\
                         & 336      & 0.246 & \textbf{0.280}                 & \textbf{0.245} & 0.282                    & 0.249 & 0.283                    & 0.339                    & 0.377 & 0.351 & 0.391                   & 0.583 & 0.543                 & 0.739 & 0.753                   \\
                         & 720      & 0.318 & \textbf{0.333}                 & \textbf{0.314} & 0.334                    & 0.320 & 0.335                    & 0.389                    & 0.409 & 0.415 & 0.426                   & 0.916 & 0.705                 & 1.004 & 0.934                   \\ 
\hline
\multirow{4}{*}{ETTh1}   & 96       & 0.379 & 0.404                 & \textbf{0.370} & 0.400                    & 0.375 & \textbf{0.399}                    & 0.376                    & 0.415 & 0.435 & 0.446                   & 0.941 & 0.769                 & 0.664 & 0.612                   \\
                         & 192      & \textbf{0.413} & \textbf{0.421}                 & \textbf{0.413} & 0.429                    & 0.414 & \textbf{0.421}                    & 0.423                    & 0.446 & 0.456 & 0.457                   & 1.007 & 0.786                 & 0.790 & 0.681                   \\
                         & 336      & 0.426 & \textbf{0.431}                 & \textbf{0.422} & 0.440                    & 0.431 & 0.436                    & 0.444                    & 0.462 & 0.486 & 0.487                   & 1.038 & 0.784                 & 0.891 & 0.738                   \\
                         & 720      & \textbf{0.447} & \textbf{0.463}                 & \textbf{0.447} & 0.468                    & 0.449 & 0.466                    & 0.469                    & 0.492 & 0.515 & 0.517                   & 1.144 & 0.857                 & 0.963 & 0.782                   \\ 
\hline
\multirow{4}{*}{ETTh2}   & 96       & 0.275 & \textbf{0.336}                 & \textbf{0.274} & 0.337                    & \textbf{0.274} & \textbf{0.336}                    & 0.332                    & 0.374 & 0.332 & 0.368                   & 1.549 & 0.952                 & 0.645 & 0.597                   \\
                         & 192      & \textbf{0.338} & \textbf{0.379}                 & 0.341 & 0.382                    & 0.339 & 0.379                    & 0.426                    & 0.446 & 0.426 & 0.434                   & 3.792 & 1.542                 & 0.788 & 0.683                   \\
                         & 336      & \textbf{0.327} & 0.381                 & 0.329 & 0.384                    & 0.331 & \textbf{0.380}                    & 0.477                    & 0.447 & 0.477 & 0.479                   & 4.215 & 1.642                 & 0.907 & 0.747                   \\
                         & 720      & \textbf{0.378} & \textbf{0.421}                 & 0.379 & 0.422                    & 0.379 & 0.422                    & 0.453                    & 0.469 & 0.453 & 0.490                   & 3.656 & 1.619                 & 0.963 & 0.783                   \\ 
\hline
\multirow{4}{*}{ETTm1}   & 96       & \textbf{0.288} & \textbf{0.342}                 & 0.293 & 0.346                    & 0.290 & \textbf{0.342}                    & 0.326                    & 0.390 & 0.510 & 0.492                   & 0.626 & 0.560                 & 0.543 & 0.510                   \\
                         & 192      & \textbf{0.331} & 0.370                 & 0.333 & 0.370                    & 0.332 & \textbf{0.369}                    & 0.365                    & 0.415 & 0.514 & 0.495                   & 0.725 & 0.619                 & 0.557 & 0.537                   \\
                         & 336      & \textbf{0.360} & \textbf{0.390}                 & 0.369 & 0.392                    & 0.366 & 0.392                    & 0.392                    & 0.425 & 0.510 & 0.492                   & 1.005 & 0.741                 & 0.754 & 0.655                   \\
                         & 720      & \textbf{0.416} & 0.425                 & \textbf{0.416} & \textbf{0.420}                    & 0.420 & 0.424                    & 0.446                    & 0.458 & 0.527 & 0.493                   & 1.133 & 0.845                 & 0.908 & 0.724                   \\ 
\hline
\multirow{4}{*}{ETTm2}   & 96       & \textbf{0.163} & \textbf{0.253}                 & 0.166 & 0.256                    & 0.165 & 0.255                    & 0.180                    & 0.271 & 0.205 & 0.293                   & 0.355 & 0.462                 & 0.435 & 0.507                   \\
                         & 192      & 0.221 & 0.293                 & 0.223 & 0.296                    & \textbf{0.220} & \textbf{0.292}                    & 0.252                    & 0.318 & 0.278 & 0.336                   & 0.595 & 0.586                 & 0.730 & 0.673                   \\
                         & 336      & 0.278 & 0.330                 & \textbf{0.274} & \textbf{0.329}                    & 0.278 & \textbf{0.329}                    & 0.324                    & 0.364 & 0.343 & 0.379                   & 1.270 & 0.871                 & 1.201 & 0.845                   \\
                         & 720      & 0.366 & \textbf{0.383}                 & \textbf{0.362} & 0.385                    & 0.367 & 0.385                    & 0.410                    & 0.420 & 0.414 & 0.419                   & 3.001 & 1.267                 & 3.625 & 1.451                   \\
\hline\hline
\end{tabular}
\label{tab:logtst}
\end{table*}

\subsubsection{Results} Results in Table \ref{tab:logtst} show the accuracy of our LoGTST design when compared against other industry-leading architectures while functioning with less than half of the parameters available to most other models. Comparison is made across five different datasets using both the mean square error and mean absolute error. In both error measurements, across varying lengths of time prediction and across multiple time series datasets the LoGTST consistently improves upon or nearly matches the PatchTST/64 design which proved the closest competition. LoGTST does so while having only 45\% the available parameters of PatchTST/64 and 58\% of the parameters of PatchTST/42. When examined more closely, the toughest competition is in the area of shorter-term look-ahead where the metric is held to only 96. In this case, our LoGTST was marginally outperformed by the PatchTST networks but only by MSE of less than 0.01  and only in 3/5 cases. When the time length metric was increased to a longer look-ahead of 192, 336, or 720, the LoGTST is superior in 10/15 cases and is within 0.004 or less MSE across all other five.

\subsection{Distributed Time Series Forecasting}

\begin{figure*}[tbh]
\begin{center}
\includegraphics[width=180mm]{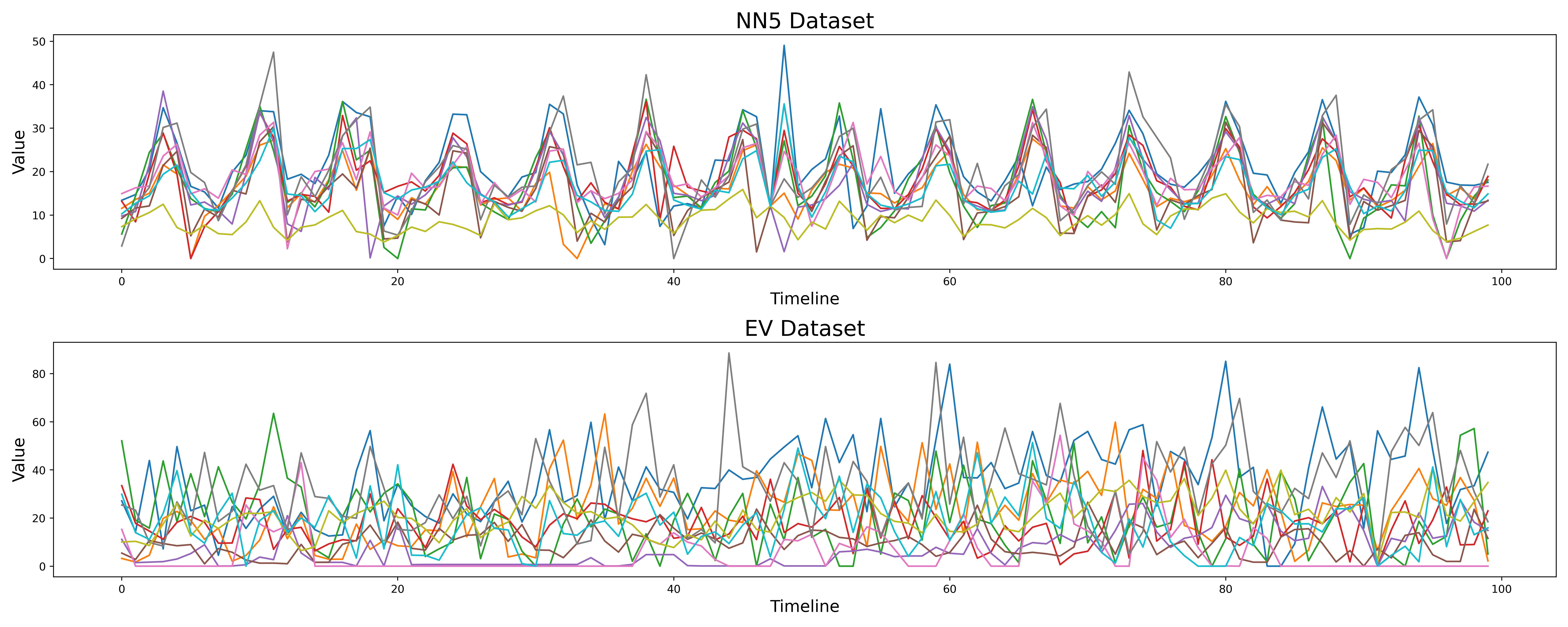}
\end{center}
\vspace*{-.2in}
\caption{Overview of the data for NN5 and EV charging datasets. We randomly select 10 clients from day 0 to day 99 of each dataset for data visualization. We can notice that the pattern of NN5 is much more obvious than EV charging dataset, where some of the data points are missing for some clients.}
\label{data_plot}
\end{figure*}

\subsubsection{Datasets} 

The EV dataset \cite{dun} was obtained
from charging stations in Dundee city, the UK
during 2017-2018, which has 65,601 transactions that
include IDs from 58 CSs, transaction ID for each CS, EV
charging date, EV charging time, and consumed energy (in
kWh) for each transaction.
There are a large amount of missing data in the dataset, so we clean the entire dataset by removing the charging stations that stopped providing data during the period. Different from previous works \cite{1ref,ho}, we merge the 24-hour energy consumption into one day and focus on daily prediction since most of the hourly data is zero and involves much more randomness compared with daily data. As shown in Figure \ref{data_plot}, even after the data cleaning, the data quality for EV dataset \cite{dun} is still quite low, which may not be able to fully benchmark the performance of current models. Thus, we also conduct experiments on NN5 dataset \cite{nn5}, which as shown in Figure \ref{data_plot} is of high quality and has a clear seasonal pattern to demonstrate the generalization ability of our method. The NN5 data consists of two years of daily cash money demand at various automatic teller machines (ATMs or cash machines) at different locations in England.

\subsubsection{Setting} Following previous work \cite{1ref,ho}, all the clients are clustered using K-means clustering algorithm based on the distances measured by dynamic time warping (DTW) \cite{muller2007dynamic}. Then, the FL process are conducted independently between different clusters. The look-back window is set to be 128 steps and the prediction horizons are scheduled to be 4 and 2 for NN5 and EV datasets considering their dataset sizes. We still use Adam optimizer with the initial learning rate to be $10^{-3}$. The ratio of selected clients is set to be 50\% for all three methods: Online-Fed, PSO-Fed, and PSGF-Fed. Both of the ratio of forward parameters of PSGF-Fed and the ratio of sharing parameters of PSO-Fed and PSGF-Fed are adjusted accordingly to find their optimal performance. The training process will be stopped when the model reaches convergence (the training loss stops decreasing for 10 rounds). RMSE is used to quantify the prediction performance.

\subsubsection{Results (NN5)}

As seen in Table II, when trained and tested against the NN5 time series dataset, the initial online FL, data-intensive training performs well with an MSE of 6.02 but at the cost of transferring approximately 1.5 billion parameters during training. The previous generation PSO FL architecture achieves slightly worse performance of 6.10 while only transferring ~0.75 billion parameters. Our PSGF FL model achieves 6.08 MSE while only transferring ~0.38 billion parameters under the 30/30 hyper-parameter implementation. This is a significant improvement in communication cost overhead with a marginal performance improvement in addition when compared to the PSO FL model.

\begin{figure}[tbh]
\begin{center}
\includegraphics[width=85mm]{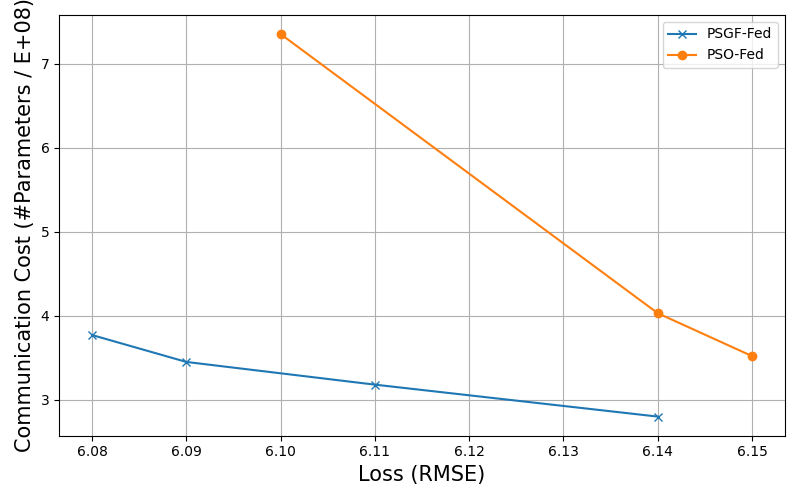}
\end{center}
\vspace*{-.2in}
\caption{The trade-off between communication cost and prediction loss (bottom left is better). With similar prediction accuracy, PSGF-Fed reduces the communication cost by at least 25\% of PSO-Fed.}
\label{trade_off}
\end{figure}

Figure \ref{trade_off} depicts the trade-off between communication cost and prediction loss. It is seen that With similar prediction accuracy, PSGF-Fed reduces the communication cost by at least 25\% of PSO-Fed for the NN5 dataset.

%1) NN5 table here:
\begin{table}[tbh]
\caption{Experimental results of different federated learning policies for NN5 dataset. The second column refers to the ratio of shared parameters between selected clients and the server. The results for the lowest communication cost and loss are folded and the results for the best trade-off are underlined.}
\centering
\begin{tabular}{lccc} 
\hline\hline
Method                         &      & \multicolumn{1}{c}{\#Params (Comm.)} & \multicolumn{1}{c}{Loss (RMSE)}  \\ 
\hline\hline
Online-Fed                     &      & 1.53E+09                             & \textbf{6.02}                             \\ 
\hline
\multirow{4}{*}{PSO-Fed \cite{gogineni2022communication}}       & 50\% & 7.35E+08                             & 6.10                             \\
                               & 40\% & 4.03E+08                             & 6.14                             \\
                               & 30\% & 3.52E+08                             & 6.15                             \\
                               & 20\% & \textbf{1.24E+08}                             & 6.28                             \\ 
\hline
\multirow{4}{*}{PSGF-Fed-20\%} & 50\% & 5.75E+08                             & 6.10                             \\
                               & 40\% & 4.21E+08                             & 6.10                             \\
                               & 30\% & \underline{3.45E+08}                             & \underline{6.09}                             \\
                               & 20\% & \underline{2.80E+08}                             & \underline{6.14}                             \\ 
\hline
\multirow{4}{*}{PSGF-Fed-30\%} & 50\% & 5.21E+08                             & 6.11                             \\
                               & 40\% & 5.25E+08                             & 6.10                             \\
                               & 30\% & \underline{3.77E+08}                             & \underline{6.08}                             \\
                               & 20\% & \underline{3.18E+08}                             & \underline{6.11}                             \\
\hline\hline
\end{tabular}
\label{tab:nn5_fed}
\end{table}

\subsubsection{Results (EV)}

Similar to the NN5 results, the PSO federate learning is able to reduce the number of parameters passed during training by approximately 50 percent while maintaining performance within 2 percent of the original online federate learning model. As shown in Table III, our PSGF architecture is demonstrated to reduce passed parameter numbers by 77 percent while maintaining performance similar to that of the PSO model. Additionally, when given a matched communication budget as the PSO model, our PSGF architecture is able to outperform the PSO by 0.25 MSE. This shows the flexibility and robustness of our improved PSGF model in data overhead as well as performance.

%2) Add UK charging table here:
\begin{table}[tbh]
\caption{Experimental results of different federated learning policies for UK EV dataset. The second column refers to the ratio of shared parameters between selected clients and the server. The results for the lowest communication cost and loss are folded and the results for the best trade-off are underlined.}
\centering
\begin{tabular}{lccc} 
\hline\hline
\multicolumn{1}{l}{Method}     & \multicolumn{1}{c}{} & \multicolumn{1}{c}{\#Params (Comm.)} & \multicolumn{1}{c}{Loss (RMSE)}  \\ 
\hline\hline
\multicolumn{1}{l}{Online-Fed} & \multicolumn{1}{c}{} & 9.07E+06                             & 10.46                            \\ 
\hline
\multirow{4}{*}{PSO-Fed \cite{gogineni2022communication}}       & 50\%                 & 4.84E+06                             & 10.68                            \\
                               & 40\%                 & 3.77E+06                             & 10.89                            \\
                               & 30\%                 & 2.75E+06                             & 10.85                            \\
                               & 20\%                 & 2.11E+06                             & 11.14                            \\ 
\hline
\multirow{4}{*}{PSGF-Fed-20\%} & 50\%                 & \underline{4.82E+06}                             & \underline{\textbf{10.43}}                            \\
                               & 40\%                 & 4.28E+06                             & 10.54                            \\
                               & 30\%                 & 2.96E+06                             & 10.67                            \\
                               & 20\%                 & 2.11E+06                             & 10.64                            \\ 
\hline
\multirow{4}{*}{PSGF-Fed-30\%} & 50\%                 & 4.46E+06                             & 10.64                            \\
                               & 40\%                 & 4.18E+06                             & 10.63                            \\
                               & 30\%                 & 3.28E+06                             & 10.65                            \\
                               & 20\%                 & \underline{\textbf{2.08E+06}}                             & \underline{10.60}                            \\
\hline\hline
\end{tabular}
\label{tab:uk_fed}
\end{table}

\section{Conclusions}

In this paper, we have proposed a communication-efficient time series forecasting model by combining LoGTST in transformer design and a PSGF-Fed strategy in FL.
Simulations have demonstrated that our LoGTST trained with PSGF-Fed offers the best trade-off between communication overhead and
performance for energy demand forecasting of a geographically dispersed series of EV charging stations.
Tests of our proposed forecasting model on other time series datasets also showed superior performance over existing approaches, making it applicable to generalized time series prediction beyond energy demand for EV charging.

\section{Acknowledgements}

Z. Xiong would like to acknowledge fruitful discussions with  Michail Gkagkos.

%\newpage
\printbibliography

\end{document}